# SEA: A COMBINED MODEL FOR HEAT DEMAND PREDICTION

**Jiyang Xie[1], Jiaxin Guo[1], Zhanyu Ma[1,*], Jing-Hao Xue[2], Qie Sun[3], Hailong Li[4,5], and Jun Guo[1]**

[1]Pattern Recognition and Intelligent Systems Lab., Beijing University of Posts and Telecommunications, Beijing 100876, China
[2]Department of Statistical Science, University College London, London, United Kingdom
[3]Institute of Thermal Science and Technology, Shandong University, Jinan 250100, China
[4]School of Business, Society and Engineering, Mälardalen University, Västerås, Sweden
[5]Tianjin Key Laboratory of Refrigeration Technology, School of Mechanical Engineering, Tianjin University of Commerce, Tianjin 300134, China
{xiejiyang2013, guojiaxin, mazhanyu, guojun}@bupt.edu.cn, jinghao.xue@ucl.ac.uk, qie@sdu.edu.cn, hailong.li@mdh.se

**Abstract:** Heat demand prediction is a prominent research topic in the area of intelligent energy networks. It has been well recognized that periodicity is one of the important characteristics of heat demand. Seasonal-trend decomposition based on LOESS (STL) algorithm can analyze the periodicity of a heat demand series, and decompose the series into seasonal and trend components. Then, predicting the seasonal and trend components respectively, and combining their predictions together as the heat demand prediction is a possible way to predict heat demand. In this paper, STL-ENN-ARIMA (SEA), a combined model, was proposed based on the combination of the Elman neural network (ENN) and the autoregressive integrated moving average (ARIMA) model, which are commonly applied to heat demand prediction. ENN and ARIMA are used to predict seasonal and trend components, respectively. Experimental results demonstrate that the proposed SEA model has a promising performance.

## 1 Introduction

Intelligent energy networks (IENs) are networks which intelligently optimize energy supply based on sharing information bi-directionally between producers and consumers [1]. Energy big data generated in the IENs provide a basis to predict the future energy consumption and determine an energy profile [2]. The big data on energy demands have enable utility companies to optimize their operation plans in order to save production cost, increase benefit and improve the robustness of energy supply.

Centralized district heating systems have been widely adopted to provide space heating in urban areas due to the advantages of, for example, high efficient and environment friendly [3]. Accurate heat demand prediction is of great significance to district heating in order to optimize production plan to reduce the cost [4].

Energy demand prediction has been intensively studied, and various methods have been proposed. Generally, the prediction is based on time series analysis. In recent years, linear regression models were developed to analyze the regularity of the time series, especially for energy demand [5, 6, 7]. there are also many statistical models that use nonlinear fitting to map history demands and predictions by using, for example, Gaussian mixture model (GMM) [8, 9, 10], support vector machine (SVM) [11], feedforward neural network (FF-NN) [12], and Elman neural network (ENN) [3, 13, 14].

Additionally, combined models have been used to further improve the accuracy of prediction. Jovanović et al. [15] combined FF-NN, radial basis function network (RBFN), and adaptive neuro-fuzzy interference system (ANFIS) to predict heat demand. Desouky et al. [16] built a hybrid ARIMA+ANN network to improve the heat demand prediction performance. Meanwhile, based on locally weighted regression (LOESS), seasonal-trend decomposition based on LOESS (STL) has been proven as a useful algorithm to handle time series seasonal decomposition problems [17]. Grmanová et al. [18] proposed two incremental ensemble learning model, called STL+ARIMA and STL+Holt-Winters exponential smoothing (EXP), for electricity load prediction.

In previous work [3], an ENN was used to predict hourly heat demand, which showed a better performance than GMM, FF-NN, and non-linear autoregressive with exogenous inputs (NARX) neural network. However, existing machine learning models remain difficult to accurately adapt the rapid changes of hourly heat demand. In order to further improve the accuracy, we aim to combine different models to exploit the best of their worlds. Hence in this paper, a combined model, STL-ENN-ARIMA (SEA), is proposed based on STL, ENN, and ARIMA. In our SEA, after decomposing the heat demand into seasonal and trend components by STL, the seasonal components are predicted by ENNs for their strength on nonlinear time series modeling, and

---

the trend component is modeled by ARIMA for its strength and simplicity in linear time series modeling.

## 2 Model description

### 2.1 The STL-ENN-ARIMA (SEA) Model

To consider better the impacts of seasonal components with different periods of the heat demand data, the proposed SEA model combines STL, ENN, and ARIMA for heat demand prediction. Two structures of the SEA model are proposed, as shown in Figure 1, where *np* is the period parameter, which means that quasi-period of the obtained seasonal component is *np*.

As well known, a time series *Y*, for example, heat and electricity demand, can be decomposed into three components as

$$Y = T + S + R, \qquad (1)$$

where *T*, *S*, and *R* are the trend, seasonal, and remainder components, respectively. The SEA model firstly divides the heat demand into the aforementioned components by using the STL algorithm. According to the results of STL, only a seasonal component of single period has been obtained with the period parameter *np*, while the resulting trend component can also have some seasonal components. Therefore, we need to repeatedly decompose the resulting trend component to obtain different seasonal components with different values of *np*, as implemented in Figure 1.

Then, after obtaining $S_{np}, np = 3,4,12,24$ and $T = T_{24}$ from the STL algorithm in Figure 1, ENN and ARIMA are used to predict $S_{np}$ and $T$, respectively. According to [3], weather conditions, for example, ambient temperature, solar radiance, and wind speed, are also impactful factors on heat demand prediction. Therefore, we added these factors as $I$ when predicting $S_{np}$. Moreover, heat demand data in the past four hours are also used to predict the current $S_{np}$ [3]. Finally, the prediction P is equal to the sum of the predictions of $S_{np}$ and $T$.

Meanwhile, as shown in Figure 1, we proposed two possible structures of the SEA model to compare the performance of predicting $S_{np}, np = 3,4,12,24$ by 4 different ENNs respectively named structure A in Figure 1(a) and 1 ENN named structure B in Figure 1(b).

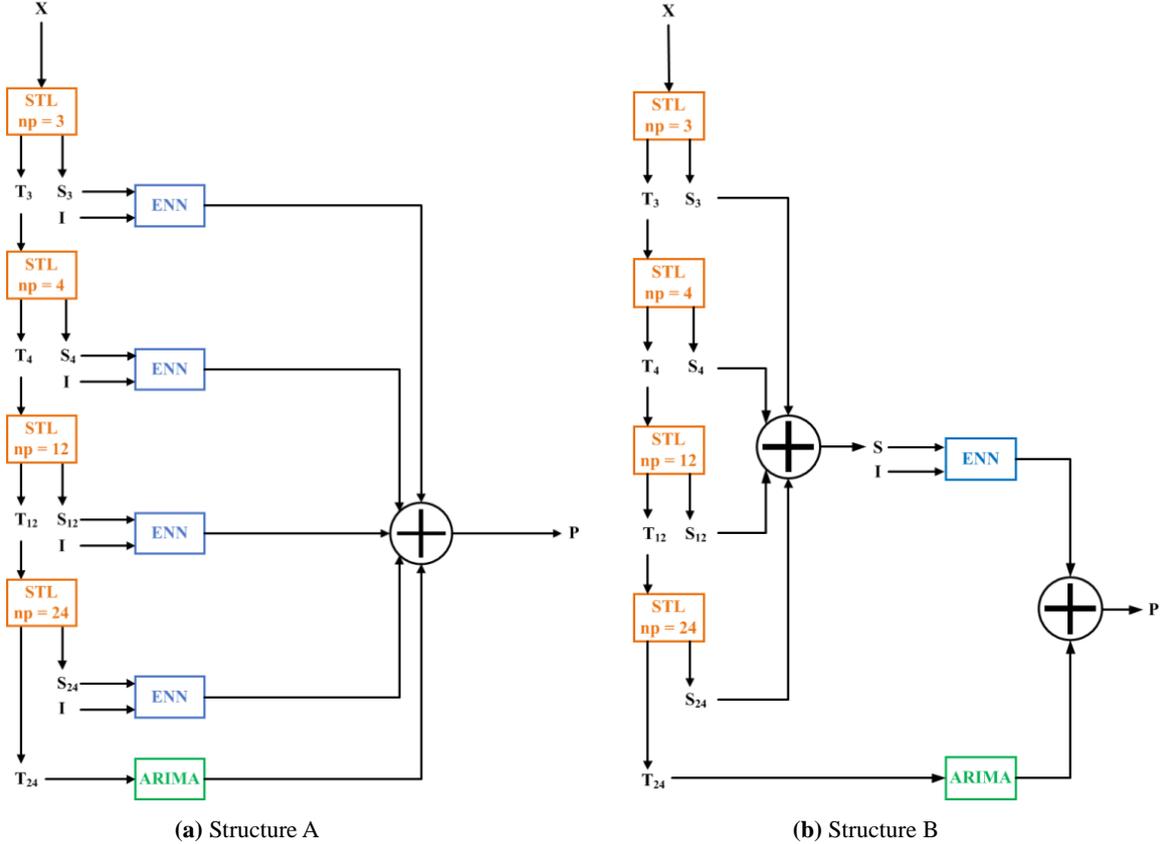

(a) Structure A  (b) Structure B

**Figure 1** Two structures of the SEA model.

### 2.2 Data Description

In the following experiments, hourly measured heat demand data (in MW) collected by a utility company during 2008-2010 (26,304 hours) including heat demand, ambient temperature, solar radiation, and wind speed are used as the training set, and the data in 2011 (8,760 hours) are used as the test set. Before training the models, we normalize all training data $\boldsymbol{x} = [x_1, x_2, \dots, x_n]$ as

$$\hat{x}_i = \frac{x_i - \mu}{\sigma}, i = 1,2,\dots,n, \qquad (2)$$

where $\hat{x}_i$ is the element of the normalized data

$\hat{x} = [\hat{x}_1, \hat{x}_2, ..., \hat{x}_n]$, $\mu$ and $\sigma$ are the mean and standard deviation of $x$, and $n$ is the number of elements in $x$ and $\hat{x}$.

Figure 2 shows the actual heat demand data from January 1st to February 29th in 2008 (1,440 hours), and the corresponding seasonal and trend components decomposed by using STL. This trend component describes overall trend of all data, and the seasonal components describe different periodic components.

### 2.3 Performance Indicators

To evaluate the performance of the heat demand prediction models, mean average percentage error (MAPE) and root mean square error (RMSE) are used as the evaluation criteria. MAPE is a relative measurement of the model, defined as

$$MAPE = \frac{1}{n}\sum_{i=1}^{n}\frac{|y_i - y_{pi}|}{y_i} \times 100\%, \quad (3)$$

where $y_i$ is the actual heat demand value, $y_{pi}$ is the corresponding heat demand prediction value, and $n$ is the prediction step length. Meanwhile, RMSE is an absolute measurement of the model, defined as

$$RMSE = \sqrt{\frac{1}{n}\sum_{i=1}^{n}(y_i - y_{pi})^2}. \quad (4)$$

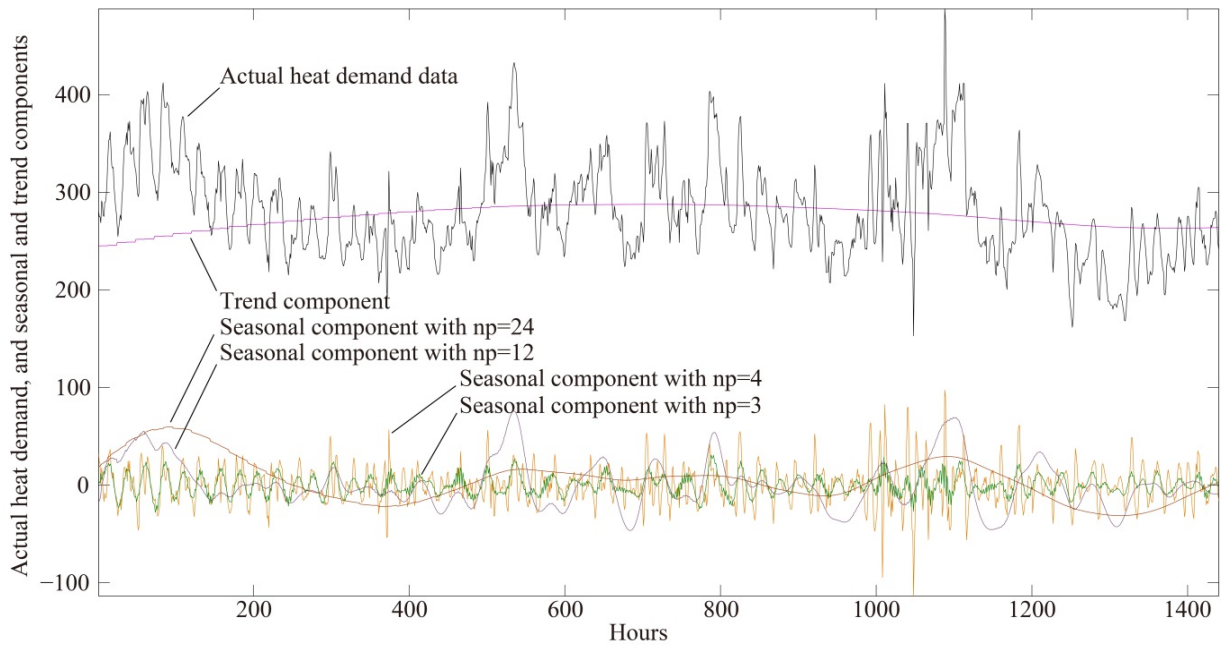

**Figure 2** Actual heat demand and seasonal and trend components from January 1st to February 29th in 2008.

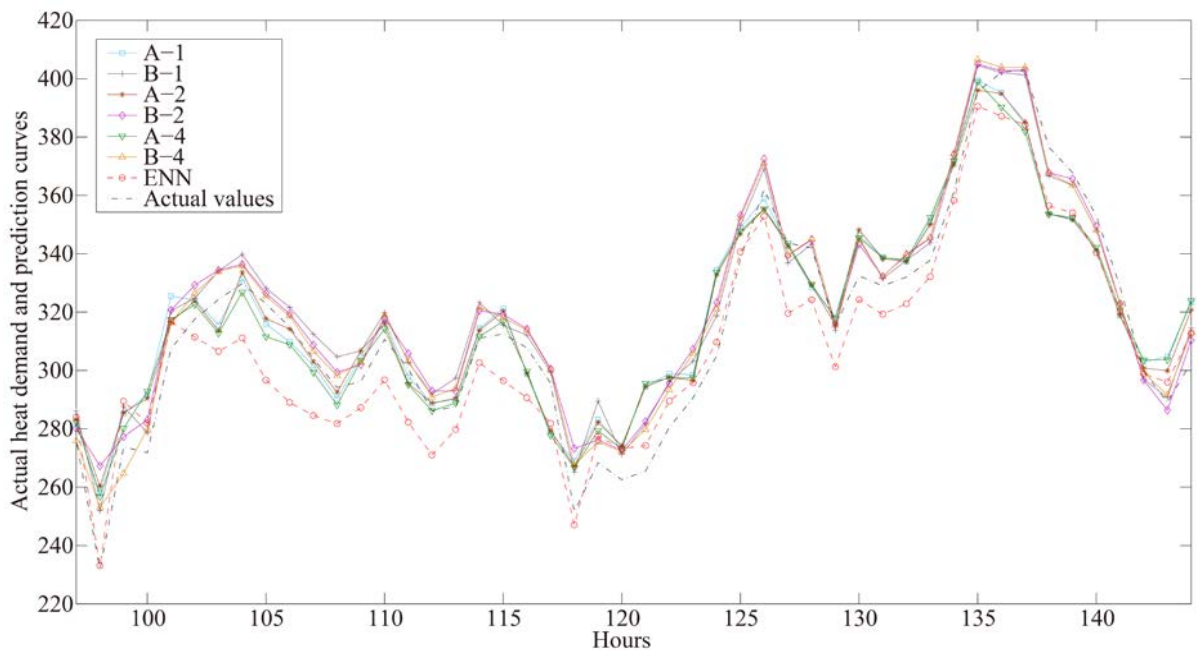

**Figure 3** Actual heat demand and the corresponding predictions using different models.

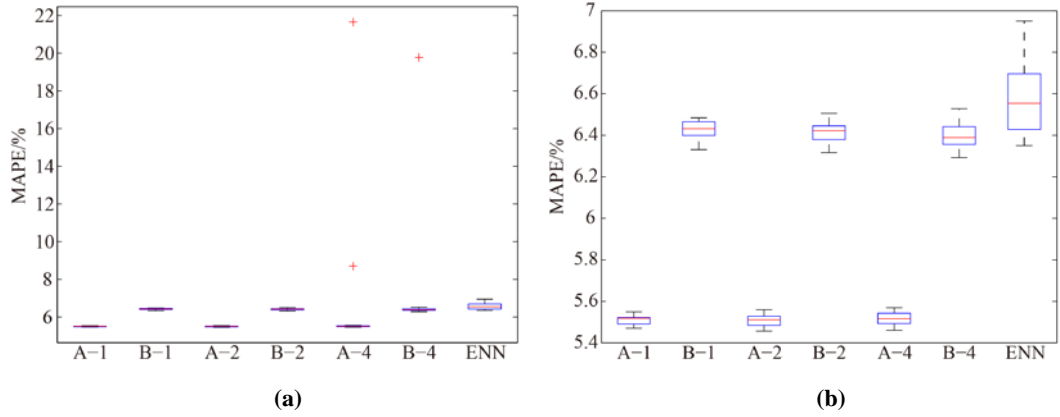

**Figure 4** **(a)** Boxplots of MAPE; **(b)** Boxplots of MAPE in the range of 5.4%~7.0%.

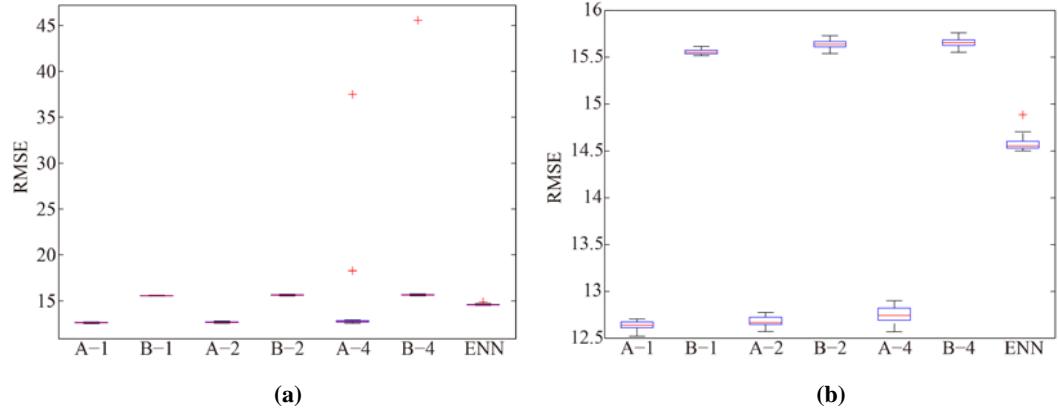

**Figure 5** **(a)** Boxplots of RMSE; **(b)** Boxplots of RMSE in the range of 12.5~16.0.

## 3 Experimental Results and Discussions

In order to investigate the performance improvement of the proposed models, heat demand predictions were implemented with the data described in Section 2.2. The ENN in models A and B have the same hidden layer node number which is set as 15. Here, 6 models are built in Table I to discuss the impact of hidden layer number for each ENN in models A and B. Meanwhile, an ENN with 8 hidden layers is created as the baseline.

**Table I** Different models for heat demand predictions.

| Hidden layer number | 1 | 2 | 4 |
|---|---|---|---|
| Model A | "A-1" | "A-2" | "A-4" |
| Model B | "B-1" | "B-2" | "B-4" |

Figure 3 compares the actual heat demand of the test set and the corresponding predictions (in MW) using different models in a short interval as an example. Generally speaking, all models are capable of reflecting the variation of heat demand. Moreover, Table II shows the comparison of different models by means and variances of both MAPE and RMSE defined in (3) and (4), respectively. The smallest mean and variance are highlighted in bold. All the models were trained for 20 times to generate the distributions of MAPE and RMSE with boxplots, which are shown Figures 4 and 5, respectively. Considering the outliers in Figures 4(a) and 5(a), we resize the figures in Figure 4(b) and 5(b) to show the boxplots more clearly.

**Table II** Comparison of different models by means and variances of MAPE and RMSE.

| Model | MAPE | | RMSE | |
|---|---|---|---|---|
| | Mean | Variance | Mean | Variance |
| A-1 | **5.51%** | **5.06E-08** | **12.63** | 2.74E-03 |
| B-1 | 6.43% | 2.14E-07 | 15.56 | **1.00E-03** |
| A-2 | **5.51%** | 6.34E-08 | 12.68 | 3.21E-03 |
| B-2 | 6.42% | 1.99E-07 | 15.64 | 2.35E-03 |
| A-4 | 6.48% | 1.39E-03 | 14.25 | 3.31E+01 |
| B-4 | 7.07% | 9.43E-04 | 17.15 | 4.70E+01 |
| ENN | 6.57% | 2.65E-06 | 14.59 | 7.71E-03 |

Based on these results, model "A-1" has the best performance on the mean and variance of MAPE, and on the mean of RMSE. Meanwhile, for MAPE, all "A" models are better than other models as shown in Figure 4(b). Similarly, for RMSE in Figure 5(b), model "A-1" has the best performance, and all "B" models are even worse than the ENN.

Tables III and IV show the *p-values* of Wilcoxon signed-rank test for comparing each pair of models in terms of MAPE and RMSE, respectively, and the statistical significance level is 0.05 ( $\alpha = 0.05$ ). According to Tables III and IV, all "A" models have statistically significant difference from all other models in both MAPE and RMSE, which indicates that model A has significant performance improvement compared with model B and ENN. Even though models "A-1", "A-2", and "A-4" are not statistically significantly different in MAPE, model "A-1" has fewer parameters and is less prone to overfitting due to the smaller number of hidden layers. Thus, model "A-1" is better than models "A-2" and "A-4".

**Table III** *P-values* of MAPE (Notation: "*": $p \leq 0.05$; "**": $p \leq 0.01$; "***": $p \leq 0.001$; and "****": $p \leq 0.0001$; otherwise $p > 0.05$ indicating non-significance).

| Model | A-1 | B-1 | A-2 | B-2 | A-4 | B-4 | ENN |
|---|---|---|---|---|---|---|---|
| A-1 |  | **** |  | **** |  | **** | **** |
| B-1 | **** |  | **** |  | * |  | ** |
| A-2 |  | **** |  | **** |  | **** | **** |
| B-2 | **** |  | **** |  | * |  | ** |
| A-4 |  | * |  | * |  | * | * |
| B-4 | **** |  | **** |  | * |  | ** |
| ENN | **** | ** | **** | ** | * | ** |  |

**Table IV** *P-values* of RMSE.

| Model | A-1 | B-1 | A-2 | B-2 | A-4 | B-4 | ENN |
|---|---|---|---|---|---|---|---|
| A-1 |  | **** | * | **** | *** | **** | **** |
| B-1 | **** |  | **** | **** | ** | *** | **** |
| A-2 | * | **** |  | **** | ** | **** | **** |
| B-2 | **** | **** | **** |  | ** |  | **** |
| A-4 | *** | ** | ** | ** |  | ** | * |
| B-4 | **** | *** | **** |  | ** |  | **** |
| ENN | **** | **** | **** | **** | * | **** |  |

## 4 Conclusions

In order to improve the performance of heat demand prediction, a combined model based on STL, ENN, and ARIMA is proposed and named as SEA. Two structures including A (*i.e.*, predicting seasonal components by 4 different ENNs respectively, and then, adding the predictions together as the seasonal components prediction) and B (*i.e.*, adding seasonal components together as the seasonal component, and predicting it by an ENN), are constructed. Compared with an ENN with 8 hidden layers, the structure A with 1-hidden-layer ENNs shows the best performance. The SEA model can not only provide accurate demand prediction, but also solve other similar nonlinear time series prediction problems. Future work will focus on investigating more effective combinations seasonal and trend components to further improve the performance.

### Acknowledgements

This work was supported in part by the National Natural Science Foundation of China (NSFC) No. 61773071, 61628301, in part by the Beijing Nova Program No. Z171100001117049, in part by the Beijing Natural Science Foundation (BNSF) No. 4162044.